\documentclass{article}
\usepackage[preprint]{neurips_2026}
\usepackage[utf8]{inputenc}
\usepackage[T1]{fontenc}
\usepackage{hyperref}
\usepackage{url}
\usepackage{booktabs}
\usepackage{amsmath,amsfonts}
\usepackage{microtype}

\title{What Counts as a Mistake?\\Annotating Recitation Events in Quran Memorization Transcripts}

\author{
  Mohamad Al Mdfaa\textsuperscript{1} \quad
  Nursultan Askarbekuly\textsuperscript{2} \quad
  Ahmed Helaly\textsuperscript{2} \quad
  Ubai Sandouk\textsuperscript{1} \quad
  Manuel Mazzara\textsuperscript{2} \\[0.6ex]
  \normalsize\textsuperscript{1}LemoniLab FZCO (\texttt{lemonilab.com})
  \quad
  \normalsize\textsuperscript{2}Innopolis University (\texttt{innopolis.ru})
}

\begin{document}
\maketitle

\begin{abstract}
Checking Quran recitation from an ASR transcript requires distinguishing unresolved mistakes from repetitions, repairs, opening formulas and accepted spelling differences. We report a completed human annotation of 100 production recording cases: 348 scored units and 162 localized events across ten combined labels. An executable evaluator scores labels and word positions together. A plain diff reaches label-aware F1 0.525 and localization F1 0.826; adapted production cleaner/alignment components reach 0.518 and 0.786, with exact-span F1 0.505 for both. Correcting the adapter's word coordinates recovers all five annotated repetition events, showing why annotation interfaces must be checked before interpreting baseline failures. In a preliminary pilot, eight single 20-minute runs across three coding agents and eight models span label-aware F1 0.143 to 0.892: seven land far above every baseline, and one collapses below the naive diff from a missing normalization step. Across the six, 970 of 972 gold-event instances draw an overlapping prediction, so what remains is not detection but convention: span extent, and the labels whose boundary is stipulated by adjudication rather than visible in the text. Seven of 162 events defeat all six same-day runs, five of them one orthographic rule, and the strongest run still misses the same ones. No run annotated before building, so the pilot measures the algorithm half of the task only.
\end{abstract}

\section{The distinction the task turns on}
Quran memorization is normally practised with a listener who corrects the reciter. An application that stands in for that listener transcribes with commercial ASR and compares the transcript against the reference. Edit distance locates differences but does not determine their interpretation. In our production data 74\% of reviewed ayah units carry no event at all, and among those that do, a substantial share are things a human listener would never mention: the reciter repeated a phrase after a breath, said a word wrongly and immediately fixed it, or used a written form differing from the Uthmani orthography without differing in what was recited. Flagging these as unresolved mistakes can give the learner misleading feedback; we do not measure the relative user cost here.

We therefore annotate transcript-against-reference differences as typed, localized \emph{events}, and we judge the text rather than the speaker: a label records what differs between transcript and reference, not whether the reciter or the ASR caused it. Unwritten vowel and tajweed errors lie outside a transcript-only rubric and outside this work. Three things follow: an annotated corpus of production recitations, a label scheme whose distinguishing move is that a repetition, a repair and an accepted spelling variant are each their own category rather than degrees of error, and a first measurement of what a coding agent does when handed the unlabelled data and asked to solve annotation and algorithm together.

\section{Related work}
Machine support for Quran recitation has mostly worked from audio: verse delimitation with speech recognition \citep{tabbal2006recitation}, tajweed-rule classification from the acoustic signal \citep{alagrami2021smartajweed}, and pretrained encoders for Quranic ASR \citep{alharbi2026comparative}, reviewed by \citet{farghaly2021asrreview}. The datasets follow the same shape: Tarteel crowd-sources and labels recitations \citep{yousfi2018tarteel}, a later corpus labels non-Arabic speakers' recitations in six categories with inter-rater agreement \citep{salameh2024quranic}, and Quran-MD aligns verse and word audio with text \citep{salman2026quranmd}. All mark a recording or a phone correct or incorrect; we label the transcript, typing and locating every difference. Pronunciation assessment scores a learner attempt against a reference \citep{witt2000goodness}, and disfluency annotation records a repetition or a repair as its own structure rather than as an error \citep{shriberg1994}; our rubric applies the second idea to an ASR transcript against a fixed reference. The coding agents we score are normally measured on software and ML engineering benchmarks \citep{jimenez2024swebench,chan2024mlebench}.

\section{Taxonomy and annotation}\label{sec:taxonomy}
\textbf{Combined labels.} An earlier pilot used two parallel fields, an event type and a benign/mistake verdict, plus an \texttt{uncertain} escape hatch. Reviewing real recordings made both untenable: the verdict was never independent of the type, and \texttt{uncertain} absorbed exactly the cases that needed a decision. The rubric now uses one combined label per event, \texttt{substitution\_mistake}, \texttt{omission\_mistake}, \texttt{insertion\_mistake}, \texttt{substitution\_corrected}, \texttt{omission\_corrected}, \texttt{repetition\_benign}, \texttt{letters\_benign}, \texttt{spelling\_benign}, \texttt{basmala\_benign} and \texttt{isti3adha\_benign}, with \texttt{clean} for a reviewed unit carrying no event. A unit reviewed and found clean is not the same as a unit not yet reviewed, and the data distinguishes them.

\textbf{Four decisions that took adjudication.} \emph{Omissions} include missing beginnings, interiors and endings, with no exemption for the final ayah of a recording: stopping early still leaves reference material absent from the transcript, and a reviewer cannot infer intent from recording position. \emph{Repeats} span the original phrase together with its extra occurrence against a single reference copy, aligning their localization with corrected events; a wrong-then-correct attempt is a corrected event, while a correct-then-incorrect restatement is a \texttt{substitution\_mistake} spanning both attempts, with the order recorded in the note. \emph{Spelling} differences removed by an agreed context-independent normalization receive no event, and only a residual form needing contextual acceptance receives \texttt{spelling\_benign}; the accepted contexts are enumerated rather than generalized, so the category cannot quietly become blanket final-letter deletion. \emph{Localization} uses verbatim words and half-open spans into the original whitespace token arrays, with an empty hypothesis span anchoring an omission and an empty reference span anchoring an insertion.

\textbf{How the annotation was produced.} One reviewer approved every record, working from the unchanged ASR transcript beside the exact reference for each ayah, and approval is per whole recording rather than per event: a recording with any unresolved question stays out of the set instead of entering it with a hedged label. That is why the rubric has no \texttt{uncertain} category and does not need one. Uncertainty is expressed by withholding approval. All 314 ayah units and 34 opening units in the set carry that approval, and every one of the 115 within-ayah events carries a written note recording which difference was the target and why the label was chosen. Those notes are what made the failure analysis in Section~\ref{sec:pilot} possible: for each event a run got wrong we can read the reasoning the run was implicitly being asked to reproduce. They stay private with the answers.

\textbf{Scale and composition.} Table~\ref{tab:labels} gives the label counts, alongside how many teaching examples each label has and how the six same-day working runs of Section~\ref{sec:pilot} handled it. All 100 selected recording cases are human-approved: 314 ayah units, 233 of them clean, and 34 opening units, over 274 ayahs in 38 surahs. Ten learner identities contribute, one of them 46 recordings. Selection was purposive, so these counts describe the annotated set; they do not estimate error prevalence or imply generalization to unseen reciters. Three further approved recordings remain outside the scored set; their transcript-only copies join the development candidates, with annotations archived privately. We verified identifiers and transcripts against both source exports and excluded punctuation-normalized duplicates and redundant clean subpassages. Gold answers and selected membership remain in an owner-held private bundle; this draft does not release a gold sample.

\begin{table}[h]\centering\footnotesize\setlength{\tabcolsep}{4pt}
\begin{tabular}{lrrrrrl}\toprule
Label & gold & teach & correct & mislab. & missed & dominant failure\\\midrule
\texttt{substitution\_mistake}   & 62 & 7 & 94\% &  0\% &  6\% & span too narrow\\
\texttt{isti3adha\_benign}       & 29 & 0 & 100\% & 0\% &  0\% & ---\\
\texttt{omission\_mistake}       & 26 & 5 & 88\% &  0\% & 12\% & read as substitution\\
\texttt{basmala\_benign}         & 18 & 0 & 88\% &  0\% & 12\% & compound opening\\
\texttt{spelling\_benign}        & 12 & 1 & 24\% & 61\% & 15\% & read as substitution\\
\texttt{repetition\_benign}      &  5 & 1 & 63\% &  0\% & 37\% & read as insertion\\
\texttt{substitution\_corrected} &  4 & 1 & 38\% &  0\% & 62\% & read as insertion\\
\texttt{letters\_benign}         &  4 & 1 & 54\% & 46\% &  0\% & read as substitution\\
\texttt{insertion\_mistake}      &  1 & 1 & 17\% &  0\% & 83\% & read as substitution\\
\texttt{omission\_corrected}     &  1 & 1 &  0\% & 67\% & 33\% & read as corrected substitution\\
\bottomrule
\end{tabular}
\caption{Gold events per label, teaching examples supplied per label, and the outcome over the six same-day working runs (162 gold events $\times$ 6 = 972 instances; 84.0\% correct, 6.1\% localized but mislabelled, 10.0\% unmatched). \emph{Correct} requires both the label and a span pair above tolerance. Reviewed clean units carry no event and are not rows here.}
\label{tab:labels}
\end{table}

\section{An executable scorecard}
Systems receive the reviewed ayah split, the ayah ids and the exact references, so detection and segmentation are supplied rather than scored; that task is a companion submission. Per unit a system returns events of the form \texttt{\{label, hyp\_span, ref\_span\}}, with half-open spans into the transcript and reference token arrays.

\textbf{Opening formulas are scored.} Each record's isti'adhah and basmala are presented as their own unit, with the formula as the transcript and an empty reference; a compound opening yields two events over one token array. Deciding that a basmala preceding the ayah is benign rather than an insertion is the judgment under test; excluding it would hide the failure mode that matters most.

\textbf{Matching.} Predictions and gold events are paired one-to-one within a unit by maximum total similarity, so a single broad prediction is credited with at most one gold event and the rest count as misses. A pair is eligible only when \emph{both} spans overlap, $\min(\mathrm{sim}_{hyp}, \mathrm{sim}_{ref}) \ge 0.50$; averaging the two, as an earlier version did, lets an exact reference span carry a match while the hypothesis span points elsewhere. Events are validated for known labels, integer endpoints, token bounds and label-specific empty spans. Invalid predictions count as false positives. Empty spans match only empty anchors, exactly or within one token; they never match selected words. Pairing ignores labels so localization can be reported separately. Units with more than seven events on either side use deterministic greedy matching.

\textbf{Scores.} The primary measure is label-aware event F1: a paired prediction is a true positive only when its label also matches, and a mislabeled pair counts once as a false positive and once as a false negative. We report micro F1 as the headline, exact-span F1 requiring identical endpoints, macro F1 across labels present in gold, and localization precision/recall/F1 with labels dropped. We also report an application-facing cost in raw event counts, $(r\,\text{missed} + \text{false flags})$ per 100 units, at $r=1$ and $r=2$, because no exchange rate between a missed mistake and an unnecessary flag is established. Neither rate ranks systems and the ordering is unchanged from 1:1 to 5:1; the rate matters only in that predicting nothing wins below roughly 0.77:1. Predicting nothing gives F1 0 and the gold annotation gives 1. Evaluator v2.1 passes synthetic coordinate, label, anchor and adapter regression tests and private gold oracle checks. Sweeping the tolerance from 0.05 to 1.00 moves baseline micro F1 by about four points and leaves exact-span F1 unchanged, since diff-based baselines emit exact spans; 0.50 is chosen to be explainable rather than tuned. One-token anchor slack changes no baseline score across 0 to 10 and is retained because the index of an omission between two tokens is genuinely ambiguous. A system predicting looser spans will depend on the tolerance more than these baselines.

\section{What existing systems do}
\begin{table}[h]\centering\footnotesize\setlength{\tabcolsep}{4pt}
\begin{tabular}{lrrrrr}\toprule
System & micro F1 & exact F1 & macro F1 & loc F1 & review cost\\\midrule
Predict nothing & 0.000 & 0.000 & 0.000 & 0.000 & 51.2 \\
Plain diff & 0.525 & 0.505 & 0.178 & 0.826 & 23.3 \\
Adapted production components & 0.518 & 0.505 & 0.284 & 0.786 & 26.7 \\
Claude Code, Fable 5.1 (pilot) & 0.892 & 0.879 & 0.702 & 0.947 & 9.2 \\
Codex, GPT-6-Astra (pilot) & 0.892 & 0.862 & 0.701 & 0.934 & 8.1 \\
Claude Code, Opus 5 (pilot) & 0.880 & 0.849 & 0.660 & 0.931 & 9.2 \\
Codex, GPT-5.6-Sol (pilot) & 0.875 & 0.863 & 0.631 & 0.938 & 9.5 \\
Claude Code, Sonnet 5 (pilot) & 0.856 & 0.825 & 0.609 & 0.912 & 9.8 \\
Antigravity, Gemini 3.6 Flash (pilot) & 0.797 & 0.773 & 0.387 & 0.878 & 14.1 \\
Antigravity, Gemini 3.1 Pro (pilot) & 0.795 & 0.677 & 0.502 & 0.873 & 10.9 \\
Antigravity, gpt-oss-120B (pilot) & 0.143 & 0.121 & 0.017 & 0.187 & 169.3 \\
\bottomrule
\end{tabular}
\caption{Evaluator v2.1: 348 units, 162 gold events. No run crashes; the gpt-oss-120B run produces 22 invalid events, discussed below. The eight pilot rows are one 20-minute development run per agent/model pair, not the preregistered comparison of Section~\ref{sec:limits}.}
\end{table}

The plain diff locates 126 of 162 gold events (recall 0.778), with localization F1 0.826. Its label-aware recall is 0.494 and F1 is 0.525. These quantities differ: localization F1 is not the percentage of gold events found. It reaches per-label F1 0.94 for omissions and 0.85 for substitutions, and zero for the other eight labels. It emits 46 insertion predictions, 34 of them on opening-text units, whereas gold has one insertion event. Compound openings contain two gold formulas but a diff can return one insertion spanning both.

The second baseline adapts production cleaner and alignment components to the new event schema; the deployed application does not natively expose it. The adapter retains original token identities through deletion, uses alignment indices and reference offsets, and maps cleaner notes to reference phrases and both attempts. Its heuristic interprets a stumble as a repair only when the following equally long surviving span matches the reference; unlocalizable removed fragments remain extra text. Production's partial-start scoring policy is retained. These choices are part of the baseline, not additional gold decisions.

This adapter predicts five repetitions and recovers all five (F1 1.0); six corrected-substitution predictions recover one of four gold events (F1 0.20). It has nonzero F1 on four labels and zero on six, and never predicts either opening-formula label, letters, contextual spelling or omission corrections. The earlier adapter lost valid repetition credit through incorrect coordinates. The remaining label gaps motivate explicit event modeling, but they do not establish that cleaning causes the errors, or measure every feature of the deployed application. Each baseline retains its own comparison normalization, an algorithm choice rather than a change to the gold rubric.

\section{A preliminary agent pilot}\label{sec:pilot}
Each of three coding agents (Codex, Antigravity, Claude Code) was given the unlabelled 127-recording training release, the annotation guide and 20 constructed teaching examples with answers, a 20-minute development budget, and no access to gold or the internet; each froze one \texttt{solution.py} that we ran, unmodified, over the private gold set. Codex ran under GPT-5.6-Sol and GPT-6-Astra; Antigravity under Gemini 3.1 Pro, Gemini 3.6 Flash and gpt-oss-120B; Claude Code under Opus 5, Sonnet 5 and, a day later, Fable 5.1, once each. Eight runs, eight models, one run per model: this is not the repeated-run comparison of Section~\ref{sec:limits}, no model, budget or seed was frozen in advance, and nothing here ranks agents or models in general. The Fable 5.1 run came a day after the other seven and reached its tools through an explicit allow-list rather than the blanket permission bypass the same-day runs used, so we keep the aggregate analyses below on the six same-day working runs and treat it as a separate contrast.

Seven of the eight reach micro F1 0.79--0.89, far above every baseline, with zero invalid predictions across all 348 units. The spread between them is smaller than the gap from either baseline, and with one run per model it should not be read as an ordering.

\subsection*{Where the six runs fail, and why}
Table~\ref{tab:labels} splits every gold event by outcome. Detection is not the residual problem: across 972 gold-event instances, 970 drew an overlapping prediction, and only two, both \texttt{spelling\_benign}, went entirely unnoticed. What the runs lose is convention. Of the 97 unmatched instances, 95 had a prediction overlapping the gold span that fell below the matching tolerance, most often because the rubric puts both attempts of a repeat or a repair in one event while the run emitted the extra text alone; that single convention accounts for the \texttt{substitution\_corrected} and \texttt{repetition\_benign} columns, where the runs read the first attempt as an insertion. The 59 mislabelled instances concentrate in \texttt{spelling\_benign} and \texttt{letters\_benign}, both read as \texttt{substitution\_mistake}.

Only seven of the 162 gold events defeated all six runs. Five are the same orthographic case, a final alif maqsura not sustained in wasl before a following definite article, which the rubric enumerates as accepted in named contexts and which all six runs called \texttt{substitution\_mistake}. One is a restart supplying a missing preposition, annotated \texttt{omission\_corrected}, which four runs read as \texttt{substitution\_corrected}; the alternative is defensible on the transcript alone. The last is a two-attempt phrase where the gold span covers both attempts and every run marked only the second. Teaching coverage does not explain the pattern. The two opening-formula labels have no teaching example at all and are handled at 100\% and 88\%, while \texttt{spelling\_benign}, which has one, is handled at 24\%. What separates them is whether the boundary is structural, as an opening formula in its own unit is, or stipulated by adjudication, as the accepted spelling contexts are. Stipulated boundaries are the part a rubric has to carry, and one worked example does not carry them.

\subsection*{How the six runs agree with each other}
Volume is not the problem either. The six runs emit 145 to 172 events against 162 in gold, and on the 233 reviewed clean units they raise three spurious mistake flags between them: one each for Sonnet 5, Gemini 3.6 Flash and Gemini 3.1 Pro, none for the other three. False positives run from 16 to 31 per run and are dominated by \texttt{substitution\_mistake} in every case. Whatever these runs get wrong, they are not flooding a learner with invented mistakes, which is the failure mode the application can least afford.

\begin{table}[h]\centering\footnotesize\setlength{\tabcolsep}{5pt}
\begin{tabular}{lrrr}\toprule
Run & events predicted & false positives & clean units flagged\\\midrule
Codex, GPT-6-Astra & 172 & 23 & 0\\
Claude Code, Opus 5 & 156 & 16 & 0\\
Codex, GPT-5.6-Sol & 158 & 18 & 0\\
Claude Code, Sonnet 5 & 158 & 21 & 1\\
Antigravity, Gemini 3.6 Flash & 159 & 31 & 1\\
Antigravity, Gemini 3.1 Pro & 145 & 23 & 1\\
\bottomrule
\end{tabular}
\caption{Error structure of the six same-day working runs against 162 gold events on 348 units, 233 of them clean. Every run is calibrated on volume to within eleven percent of the gold count.}
\end{table}

Treating the six runs as independent readers of the same rubric gives a second view of the annotation. On 114 of the 162 gold events (70\%) all six agree with the gold label. On 43 (27\%) they split. On five (3\%) all six agree with each other and disagree with gold: four are the \texttt{spelling\_benign} wasl case read as \texttt{substitution\_mistake}, and one is the two-attempt phrase every run localized too narrowly. Unanimous disagreement is therefore not noise. It landed on one rubric rule and one span convention, and the rubric's own notes describe that rule as an approved extension of an earlier allowance rather than something a reader could derive. Consensus across independently developed solutions is a cheap signal for where a rubric is doing stipulative work, though it cannot decide whether the stipulation is right.

\textbf{The later run narrows the residue.} Fable 5.1, run a day later under the same inputs and budget, ties the best micro F1 at 0.892 and takes the best exact-span F1 (0.879), localization (0.947) and macro F1 (0.702) in the set, finishing in four minutes of its twenty. It differs from the six mainly by getting the span conventions right: \texttt{substitution\_corrected} 4 of 4, \texttt{repetition\_benign} 5 of 5 and \texttt{letters\_benign} 4 of 4, against 38\%, 63\% and 54\% correct across the six, with mean span IoU 0.995 and no clean unit flagged. What is left is almost entirely the stipulated boundary: eight of its nine mislabelled events are \texttt{spelling\_benign}, at precision 1.00 and recall 0.25, and they include the same wasl forms and the same \texttt{omission\_corrected} restart that defeated all six. One run cannot separate the model from the day or the permission setting, but it is consistent with the split above, in which convention errors yield to capability and adjudicated boundaries do not.

The failing run is informative in the other direction. gpt-oss-120B's solution never normalizes Arabic orthography before diffing, so ordinary hamza and alif-form differences register as \texttt{substitution\_mistake} on nearly every token (545 of 567 predicted events carry that one label); it also treats every opening unit as clean and produces 22 invalid events. At micro F1 0.143 it is worse than either baseline: a diff-shaped answer under the same instructions, guide and examples as the seven successful runs can still fail on the comparison step that Section~\ref{sec:taxonomy}'s rubric assumes is solved.

\section{Discussion}
\textbf{What the label scheme buys.} The scheme's distinguishing move is that a repetition, a repair and an accepted spelling variant are categories in their own right rather than milder mistakes. That move is not cosmetic on this data. Of the 162 located events, 68 are benign and 5 are corrected, so 73 of them, 45\%, would reach a learner as unresolved mistakes under any scheme that only has mistake categories, on top of the 233 units that carry no event at all. The distinction also survives contact with systems: the two baselines score 0.826 and 0.786 on localization but only 0.525 and 0.518 label-aware, and the gap is entirely naming, not finding. A checker that reports where the transcript differs from the reference is a solved problem. A checker that reports what a listener would have said is not.

\textbf{What the loop did, and what it did not.} Handed the unlabelled pool, the guide and 20 worked examples, seven of eight runs produced code that reaches 0.79 to 0.89 within 20 minutes, against 0.525 for a plain diff and 0.518 for adapted production components. The result is close enough to the ceiling that the interesting question is no longer whether the loop works on the algorithm but where its remaining errors come from, and Section~\ref{sec:pilot} answers that: convention rather than detection. The runs are also conservative in the way the application needs, raising three spurious mistake flags across 1{,}398 clean-unit decisions.

The annotation half is a different story, and the honest answer is that this pilot did not test it. Seven of the eight runs went straight to rules and never opened the training pool at all. The eighth, Fable 5.1, ran its finished detector across all 888 units, wrote the labels to a file and used their distribution and sampled spans as a development check, which is the pool used as a signal rather than annotation done first and built upon. A 20-minute budget makes the other seven rational, so nothing here says an agent will not annotate under a budget that rewards it. It does say that offering annotation as an available strategy is not enough to observe it, and that a design meaning to study the annotate-then-develop schema has to require delivery of the annotations and give the loop time to make producing them worthwhile.

\textbf{Where the two halves meet.} The failure analysis suggests the two problems are not independent. Every label the runs handle well has a boundary visible in the input: an opening formula occupies its own unit, missing reference material is missing. Every label they handle badly has a boundary fixed by adjudication: which spelling differences are accepted in which contexts, how far a span reaches when a reciter attempts a phrase twice. An agent asked to annotate would face exactly the boundaries its algorithm gets wrong, and the five events where all six runs agree against gold point at the same places. That makes agent consensus useful as a rubric audit before it is useful as a labelling shortcut: it says nothing about which reading is correct, but it locates the clauses where the rubric is stipulating rather than describing, and those are the clauses a human has to own.

\section{Limitations and what comes next}\label{sec:limits}
The annotated set is purposive and small, one learner contributes 46 of the 100 recordings, and the label distribution is heavily skewed, so macro F1 here is a coverage check, not a stable estimate. A single reviewer approved every record, so we report no inter-annotator agreement and cannot separate the rubric from that reviewer's reading of it. The written notes make each decision auditable and the failure analysis shows where independent readers would diverge, but neither substitutes for a second annotator on a sample, which is the first thing we would add. The five events where all six runs agree against gold are the natural place to start. More consequentially, all 100 selected transcripts already appear as inputs in a public release of the same recordings, and twelve in an earlier machine-labeled pilot. The adjudications and the selection membership are private, but the inputs are not, and runtime isolation cannot establish absence of prior exposure; an unseen-input claim requires a separate corpus. The prepared development release contains 127 cases and 888 units, including 44 of 63 newer recordings. It permits shared clean ayahs and different error variants, but excludes complete gold copies and event-bearing gold chunks. The one local ASR repair appears non-Quranic and is excluded. Remaining new recordings are not a frozen hidden test.

We plan to compare agents under the same unlabelled training data, rubric, examples, tools and budget. Agents may annotate, train models or write rules; annotation is an available strategy, not the experimental contrast. In the pilot, no run took it as a first step: every workspace we recovered contains rule-based code and a self-test against the supplied examples, and only the Fable 5.1 run touched the 888-unit training pool at all, labelling it with its own finished detector as a check rather than to learn from. Under a 20-minute budget that is a rational choice, and it means this pilot measures the algorithm half of the task only. Whether an agent will annotate at all, and whether its labels survive review, needs a budget that makes annotating worth the time and a delivery requirement that makes it observable. The pilot above is a first, informal instance of this design, run to check that the harness and isolation work end to end before the preregistered protocol is fixed. Isolation differed by runtime. Codex and Claude Code each ran non-interactively in a fresh git-free directory outside this repository, under their sandbox's default network restriction, with an isolated credentials-only home directory carrying no prior session history, cache or goals, one fresh instance per model. Antigravity ran in a fresh window and directory per model; its network access was not independently instrumented, and its selected model is taken on the operator's report rather than verified externally. All seven saw only the unlabelled release, the guide and the teaching examples, never gold, and froze their solution before any gold scoring. A high score supports the combined process, not each training label: code may ignore or correct its labels. Generated annotations need independent review, but a fully human-labelled training pool is unnecessary. Gold answers, selected test bundles and review history stay outside development environments; shared clean inputs and prior public exposure are disclosed. The repeated-run comparison, with budgets, models and annotation delivery fixed in advance, remains future work.
\bibliographystyle{plainnat}
\bibliography{references}

@inproceedings{tabbal2006recitation,
  title={Analysis and implementation of a" Quranic" verses delimitation system in audio files using speech recognition techniques},
  author={Tabbal, Hassan and El Falou, W and Monla, B},
  booktitle={2006 2nd international conference on information \& communication technologies},
  volume={2},
  pages={2979--2984},
  year={2006},
  organization={IEEE}
}

@article{alagrami2021smartajweed,
  title={Smartajweed automatic recognition of Arabic quranic recitation rules},
  author={Alagrami, Ali M and Eljazzar, Maged M},
  journal={arXiv preprint arXiv:2101.04200},
  year={2020}
}

@article{alharbi2026comparative,
  title={A Comparative Study of Pretrained Transformer Models for Quranic ASR: Speech Representations, Label Formats, and Dataset Composition},
  author={Hossain, Nabil Mosharraf and Islam, Riasat and Obaidellah, Unaizah},
  journal={arXiv preprint arXiv:2606.19747},
  year={2026}
}

@article{farghaly2021asrreview,
  title={Automatic speech recognition (ASR) systems for learning Arabic language and Al-quran recitation: a Review},
  author={Balula, Nazik O’mar and Rashwan, Mohsen and Abdou, Shrief},
  journal={International Journal of Computer Science and Mobile Computing},
  volume={10},
  number={7},
  pages={91--100},
  year={2021}
}

@article{yousfi2018tarteel,
  title={The Tarteel dataset: crowd-sourced and labeled Quranic recitation},
  author={Khan, Hamzah I and Abid, Abubakar and Moussa, Mohamed Medhat and Abou-Allaban, Anas},
  year={2021}
}

@article{salameh2024quranic,
  title={Quranic audio dataset: Crowdsourced and labeled recitation from non-Arabic speakers},
  author={Salameh, Raghad and Al Mdfaa, Mohamad and Askarbekuly, Nursultan and Mazzara, Manuel},
  journal={Procedia Computer Science},
  volume={246},
  pages={2684--2693},
  year={2024},
  publisher={Elsevier}
}

@inproceedings{
salman2026quranmd,
title={Quran-{MD}: A Fine-Grained Multimodal Dataset of the Quran},
author={Muhammad Umar Salman and Mohammad Areeb Qazi and Mohammed Talha Alam},
booktitle={5th Muslims in ML Workshop co-located with NeurIPS 2025},
year={2025},
}

@article{witt2000goodness,
  title={Phone-level pronunciation scoring and assessment for interactive language learning},
  author={Witt, Silke M and Young, Steve J},
  journal={Speech communication},
  volume={30},
  number={2-3},
  pages={95--108},
  year={2000},
  publisher={Elsevier}
}

@article{shriberg1994,
  title={Preliminaries to a theory of speech disfluencies},
  author={Shriberg, Elizabeth Ellen},
  journal={Doctoral dissertation, University of California at Berkeley},
  year={1994}
}

@inproceedings{jimenez2024swebench,
  title={Swe-bench: Can language models resolve real-world github issues?},
  author={Jimenez, Carlos E and Yang, John and Wettig, Alexander and Yao, Shunyu and Pei, Kexin and Press, Ofir and Narasimhan, Karthik},
  booktitle={International Conference on Learning Representations},
  volume={2024},
  pages={54107--54157},
  year={2024}
}

@inproceedings{chan2024mlebench,
  title={Mle-bench: Evaluating machine learning agents on machine learning engineering},
  author={Chan, Jun Shern and Chowdhury, Neil and Jaffe, Oliver and Aung, James and Sherburn, Dane and Mays, Evan and Starace, Giulio and Liu, Kevin and Maksin, Leon and Patwardhan, Tejal and others},
  booktitle={International Conference on Learning Representations},
  volume={2025},
  pages={50466--50494},
  year={2025}
}

\end{document}